# Flying robots for a smarter life

**Dr. Abdullatif BABA**

Kuwait College of Science and Technology (KCST), Computer Science and Engineering Department, Kuwait.
University of Turkish Aeronautical Association (UTAA), Computer Engineering Department, Ankara, Türkiye.
a.baba@kcst.edu.kw; ababa@thk.edu.tr

## Abstract

Innovative ideas are continually emerging to create better living conditions where essential human needs are fulfilled in ideal scenarios, leading us to propose modern strategies that shape the future of smart cities. In this context, flying robots are being increasingly utilized in various fields to enhance the quality of our lives. This paper illustrates new designs of flying robots that could be used to perform a variety of advanced missions, like investigating the state of high-power lines and manipulating cabling maintenance procedures when failures are detected, evaluating the state of the outer edge of sidewalks to color their partially or wholly erased parts, and spraying pesticides on trees or crops that are affected by different diseases. Creating such smart devices demands developing many other partial designs relying on AI-based algorithms, computer vision techniques, and embedded systems. A variety of techniques that we have recently developed in this field are presented here.

## 1 Introduction

The last few years were characterized by accelerated development and flexible adaptation of several unmanned aerial designs, including fixed-wing drones and multicopters, for serving as flying robots that were widely exploited to perform a variety of intelligent missions like real-time surveillance [1], emergency and rescue [2], firefighting [3][4], goods deliverance [5], and many other applications. This new technology has evident social and economic impacts on our lifestyle and represents one of the most vital future trends [16]. In this context, Suab et al. have generally explained how different UAVs could be employed in forests and farming [7]. On the other hand, the mobility and flexibility of UAVs as well as their cheap costs represent their most powerful features for developing a new generation of cellular communication flying terminals as in [8][9][10], acquiring images rich of details for executing several geological applications as suggested in [11], or monitoring and tracking wildlife features [12][18].

Many engineers and researchers look at commercial multicopters as practical devices that could be adapted to perform a wide range of applications, even though they are unable to carry considerable payloads, and their batteries may sustain up to two successive flying hours if their maximum weight, including their payload, is around ten kilos. In such a case, building advanced types of flying robots that satisfy any special requirements regarding their payload or their maximum flying range requires new, creative ideas. This paper presents a few novel designs with many technical details, including computer vision systems and some modern AI-based algorithms that are required to perform smart missions. The organization of this paper is as follows: first, we present several recently designed and presented platforms for serving in different areas.

The technical characteristics of the presented flying robots, including sensors and actuators, will be illustrated. Smart decision-making techniques using computer vision systems are detailed, whereas classical tools and ANN-based methods are deeply discussed. Finally, the paper ends with a proper conclusion.

## 2 Different designs of flying robots

Flying robots are the fruit of the scientific revolution that matured during the last century. Most of what was once a fantasy has become true, and the future still holds a lot; these flying devices are operating around us and will be used everywhere. In this context, we have recently suggested a new design of a flying robot that could be regarded as a part of a self-healing strategy for the new generation of smart grids. It will investigate the state of their transformers, isolators, control units, high-power cables, or any of their essential element [13]. The given design illustrated in Figure (1) is a quadcopter equipped with two robot arms to implement some maintenance procedures like cable jointing and replacing failed elements, as well as two grips that are used to guarantee perfect hovering stabilization when performing the previously mentioned procedures.

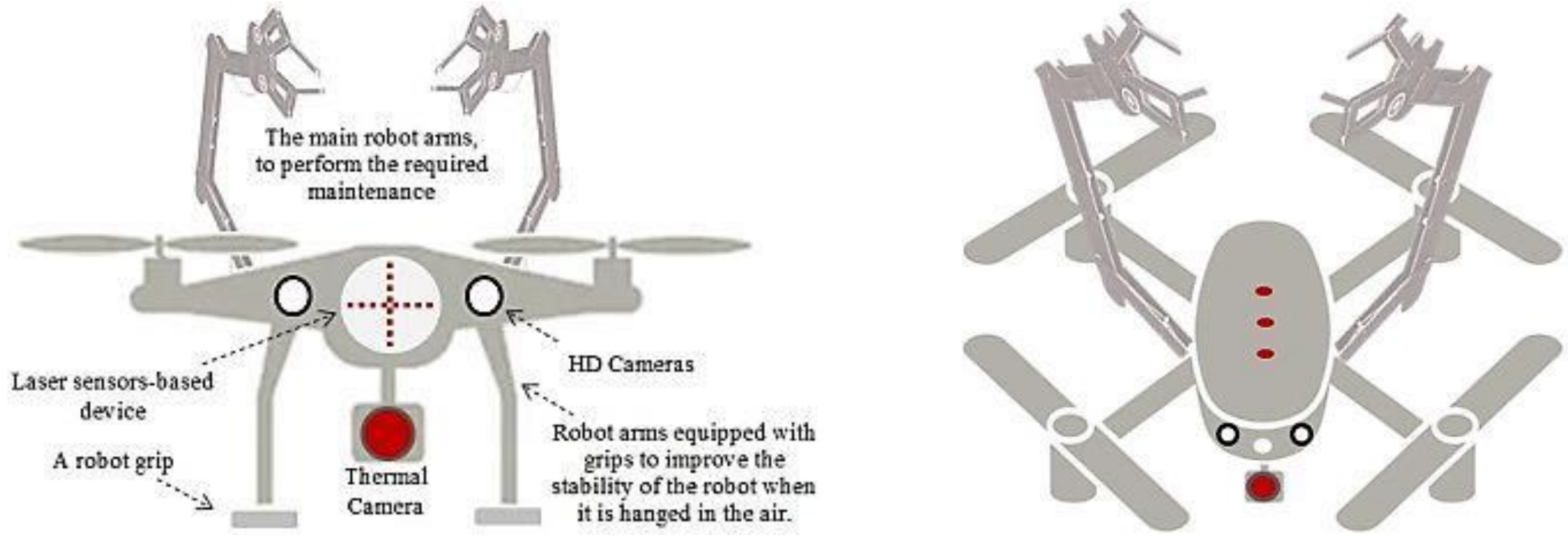

Figure 1. A front and top view of the suggested flying robot.

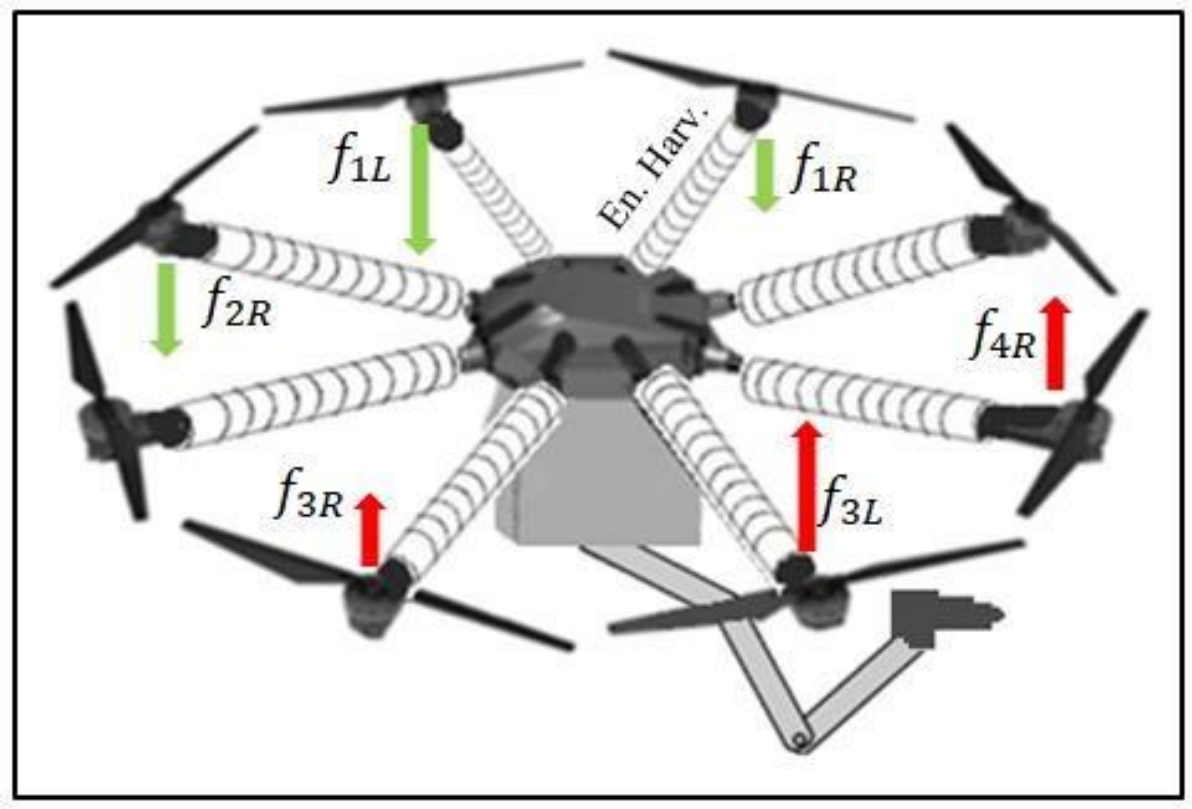

Figure 2. An octocopter provided with 8 embedded energy harvesters, and the robot arm.

Another flying robot was recently designed and will appear quickly to serve as a liquid transporter. This advanced platform is significantly needed for achieving different practical tasks like firefighting, spraying infected crops with the appropriate pesticide, or facade coloring implementations. As shown in Figure 2, this design is an octocopter equipped with a 10-liter tank capacity, an energy harvester installed on each rotor support, and a robot arm that can arbitrarily move in all directions (360 degrees). This motion may lead to destabilizing the flying robot. Hence, the desired perfect stability could be restored by increasing the lifting forces for rotors that overlap the motion angle of the robot arm while decreasing the lifting forces for rotors on the opposite side by the same values. This stabilization system relies on a fuzzy logic-based approach. The need to use eight energy harvesters in this design could be attributed to the fact that an octocopter is mainly regarded as an energy-consuming device that requires optimizing its energy management.

## 3 A brief description of hardware specifications

A variety of sensors, actuators, and embedded systems are required to accomplish the upper-mentioned designs as reported in Table (1) and explained here as follows:

- Colored images: Computer vision-based techniques are widely utilized for smart decision-making with new platforms of embedded systems. The better the quality of the given-colored image, the more reliable key features can be extracted, which may facilitate and accelerate the smart inferring system. This fact requires improving the hardware fabrication technologies as well as the software used to encode each pixel as a digital quantity. From a practical point of view, the last few years have witnessed the wide exploitation of deep learning-based techniques in many fields. For example, investigating the status of a failed electrical component or an infected tree by using a visual system (mono, stereo, or panoramic) requires taking successive frames from different angles and applying them to a previously trained ANN. It should be noted that all considered images may need a primitive processing step as they must have the same size and quality; sometimes we may need to stimulate and emphasize the differences among available local neighboring pixels.

- Laser sensors: Compared to colored images, the laser sensor may appear unable to give rich details of the given environment; it works as a blind person's cane. But from a practical point of view, its main benefit is its ability to provide accurate depth measurements in real-time. In such a case, a full description of a dynamic environment could be accomplished by using a sensor fusion technique between a laser scanner and a mono-vision system; this combination can replace a stereo-vision system, which is usually considered a time-consuming platform. In the first design presented in figure (1), a vertical set of laser sensors detects the vertical cables. In contrast, the horizontal set is used to keep the flying robot in parallel with the transformation power lines.

- Thermal images: The essential concept here is to create an infrared-based image by interpreting the emitted power from an object as a Stefan–Boltzmann constant times the thermodynamic temperature of the same object to the power of four, i.e., the emitted power is proportional to its temperature [13]. Long-wave or mid-wave could be detected at different spectrum ranges (8–14 μm) or (3–5 μm), respectively. For example, this type of image is consistently considered a rich and useful source for detecting several types of failures that appear in the form of an increase in the temperature of electrical elements.

- Rotor lifting thrust: To take off the land and fly, a multi-copter requires sufficient lifting thrust that should be provided by the installed rotors. In this case, the total weight of the multi-copter, including its main frame, all the hardware elements, and the maximal predicted load, should be considered when calculating the required thrust per rotor as follows:

$$T = \frac{2\,w\,s}{n}$$

($T$) is the calculated thrust, ($w$) is the total predicted weight, ($n$) is the number of rotors (4, 6, or 8 for a quadcopter, hexacopter, and octocopter respectively), and (s) is the safety factor that represents an additional efficiency margin that should be considered for getting the appropriate rotor, it is usually evaluated as 20% of the total weight ($w$)

Table 1. The specifications of robot elements.

| **The element** | **Weight (g)** | **Pieces** |
|---|---|---|
| Thermal Camera | 72 | 1 |
| Camera | 116 | 1 |
| Laser sensors | 850 | 12 |
| Robot arm | 940 | 2 |
| Robot arm | 640 | 2 |
| Drone Motor | 1038 | 4 |
| Drone frame | 12000 | 1 |
| IMU + GPS | 180 | 1 |
| Embedded system | 263 | 1 |
| Battery | 688 | 2 |
| Altimeter and force meter sensors | 100 | 5 |
| The total sum | 32019 | 32 |

## 4 Computer vision-based decision-making systems

In an embedded system, the inference of rational decisions requires reading and interpreting all available signals from different sensors describing the external environment. While the mechanism of decision-making that relies on advanced algorithmic approaches represents the most important and complicated part as it should be characterized by speed and accuracy. This section briefly describes a few classical and advanced methods developed for operating on the previously mentioned platforms.

### 4.1 Classical tools

The classical tools usually used to process digital images, including those captured by HD or thermal cameras, represent the shortest way of getting satisfactory results. The main drawback of such approaches is their inability to deliver the same reliable outcome for each new implementation with a new image containing novel features, different resolutions, or degraded quality. For example, some thresholding-based techniques may tend to automatically extract the appropriate threshold from a given image depending on widely used procedures like Otsu's method [14]. In such a case, a few successive pre-processing steps are commonly recommended to boost the dissimilarities between neighboring pixels, which may lead to unpredictable performance when looking at the output images. On the other hand, the Hough transform was utilized in real-time as a practical tool to extract the salient primitive features (lines, circles, or ellipses) to detect high-voltage transfer towers [13]. In contrast, the Gabor filter was combined with PCA (an algorithm to detect the collective components between neighboring pixels that have some spatial relationships) as an alternative method to perform the same objective.

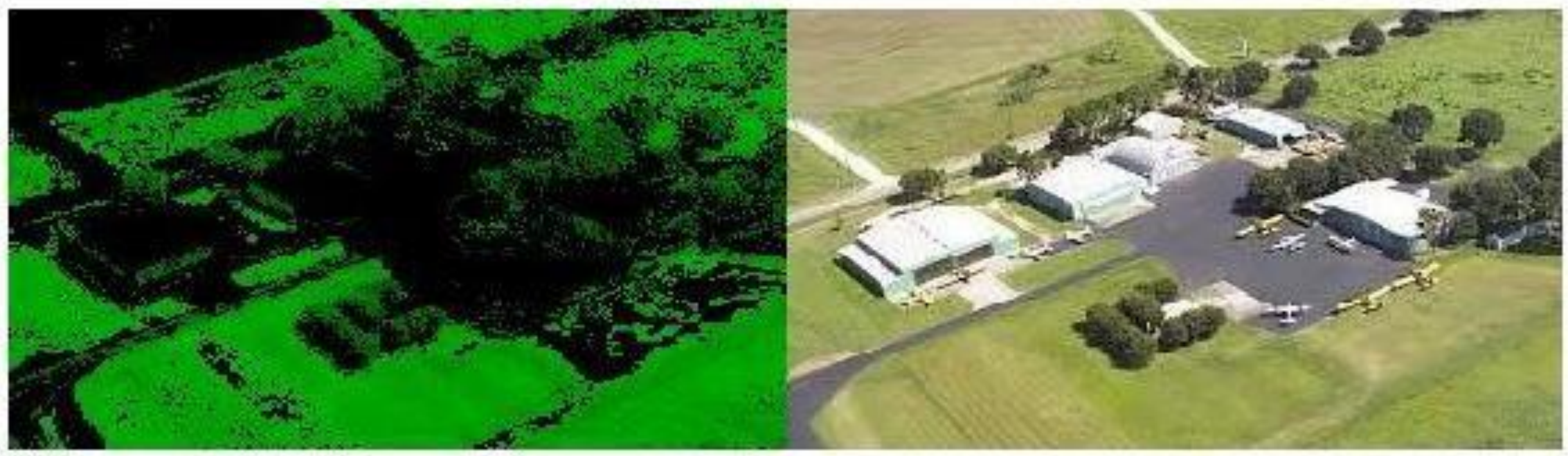

Figure 3, The total density of green space in the given image is 49.68%. The extracted image shows the distribution of green plants over the given scene.

In another study of us [15], the density of green surface was evaluated by using some classical tools of digital image processing to decide the quantity of pesticide which should be applied to trees by using an approach that relies on a fuzzy logic controller as shown in figure (3) where the total density of green space in the given image was evaluated around 49.68%. The output image shows the distribution of green plants over the given scene.

### 4.2 ANN-based tools

To overcome the disadvantages of classical methods, more advanced techniques relying on new designs of ANNs were recently proposed [16]. Broadly speaking, ANNs are used to approximate human brain performance. The raw data collected from sensors should be adapted into vectors having the same dimensions and divided into training and testing sets with a predefined split ratio. The main objective is to create an ANN able to identify the given data’s hidden characteristics and deliver the proper output when new sensorial readings are applied to its inputs. The architecture of a multilayer ANN contains an input layer where each neuron shares its own received signal with all neurons of the next layer, which could be a single hidden layer or the first of several hidden layers where most of the computational load will be done, in addition to an output layer that produces the calculated output. The backpropagation algorithm is mainly used for training ANNs. On the other hand, thanks to its main characteristic of converting any input value into a normalized output in the range [0, 1], which is suitable for probabilistic applications, the Sigmoid function, figure (4), is mostly adopted to activate (trigger) neurons.

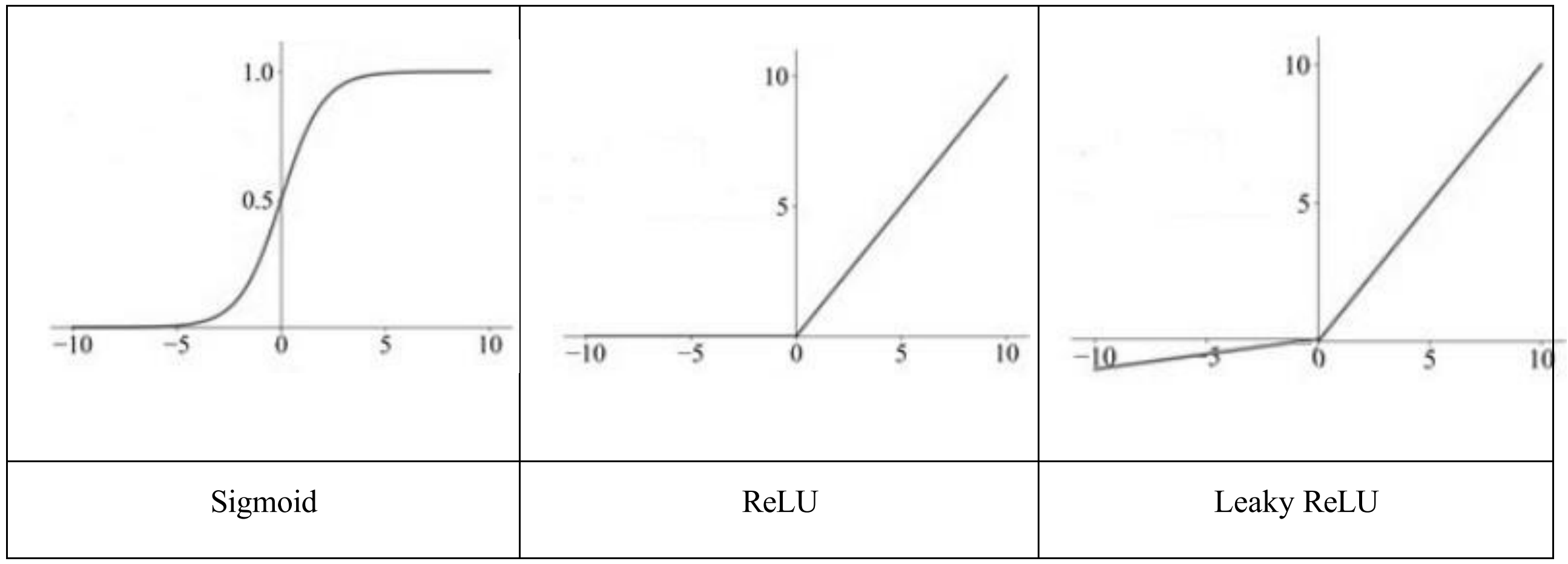

Figure 4. The three activation functions are illustrated: Sigmoid, ReLU, and Leaky ReLU.

A couple of issues that could be encountered during the training process are explained here. The first one is called the vanishing gradient problem; where in some cases, the backpropagated gradient errors become vanishingly small, consequently preventing the weights of a given neuron from changing even in small values; that means the neuron itself doesn't profit from the successive training vectors to extract the hidden features from the given data, consequently, its influence on the output becomes so tiny or negligible. In any case, we must notice that the weights of links inside an ANN are used as long-term memory compared to the biological neuron. To solve this problem, the Sigmoid activation function could be replaced with ReLU, which is faster as it does not activate all neurons simultaneously. However, when using the ReLU function, the gradient for any negative value applied to it equals zero. Accordingly, the corresponding weights will not be updated during the backpropagation phase. Such a case may lead to dead neurons with zero output in the relevant region; it should be noted that a dead neuron doesn't affect any other neuron on the next layer. To overcome this latter problem, the activation function could be modified this time by producing the Leaky ReLU, figure (4), which has a slight slope for negative values instead of a flat slope; so that negative inputs become acceptable to deliver some corresponding values on the neuron's output instead of zero. Different components of high-voltage towers were detected and classified in the same context using a Convolutional Neural Network (CNN) composed of eight layers [13]. The dataset was manually collected; composed of 120 images, a relatively small number of images. The implementation was repeated 6 times with 10-fold cross-validation, and the average accuracy of the suggested CNN is around 89%, figure (5).

Recently, we have proposed another design [17] to examine and color the outer side of partially erased sidewalks that are usually painted in two contrasting colors in a repeated way (ex., white and black, or yellow and green). In such a case, segments composed of repeated patterns (Dark, Bright, Dark) and (Bright, Dark, Bright) should be first extracted. Therefore, the wavelet transformation with the template "MexicanHat" was applied to the original image to determine the corresponding coefficient array, which is used later to build an output image as illustrated in figure (6). Then, to decide if a given block requires being plated, each pattern composed of three blocks will be converted into a vector consisting of three elements: 1 is bright, -1 is dark, and 0 is any vague level in between, figure (7). As shown in figure (8), making the correct decision requires training a Hopfield network of three neurons that memorize two fundamental vectors [1, -1, 1] and [-1, 1, -1] that will be recognized on its outputs from the first iteration when they are applied to its inputs. Controversially, applying a fully or partially incorrect vector to the inputs of the given ANN requires several successive iterations to converge toward the nearest vertex. More than 100 images were adapted for this experiment to make a dataset containing fully colored, partially erased, and fully erased sidewalks. The suggested ANN was able to classify the given image with an average accuracy of 98%.

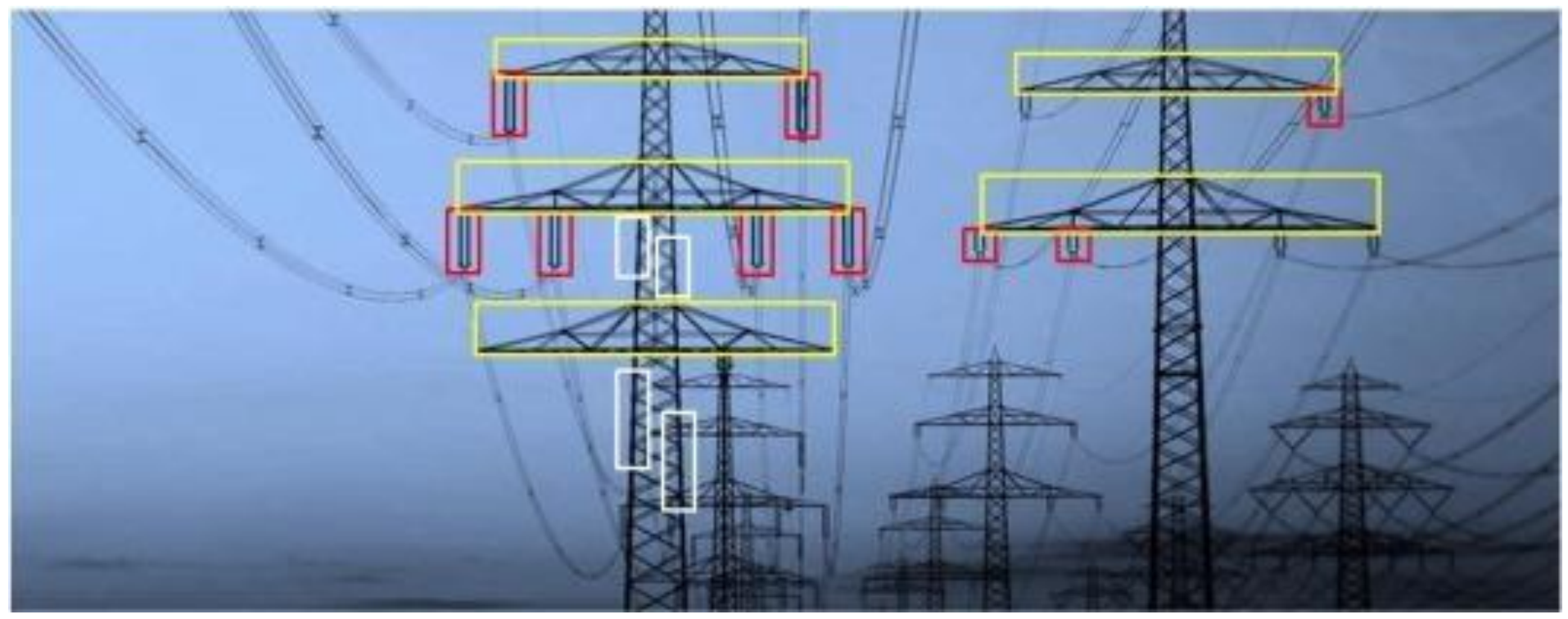

Figure 5. Different electrical components are detected. The insulators are clarified by red boxes, the triangular shapes are given inside yellow boxes, while the falsely detected features are surrounded by white boxes.

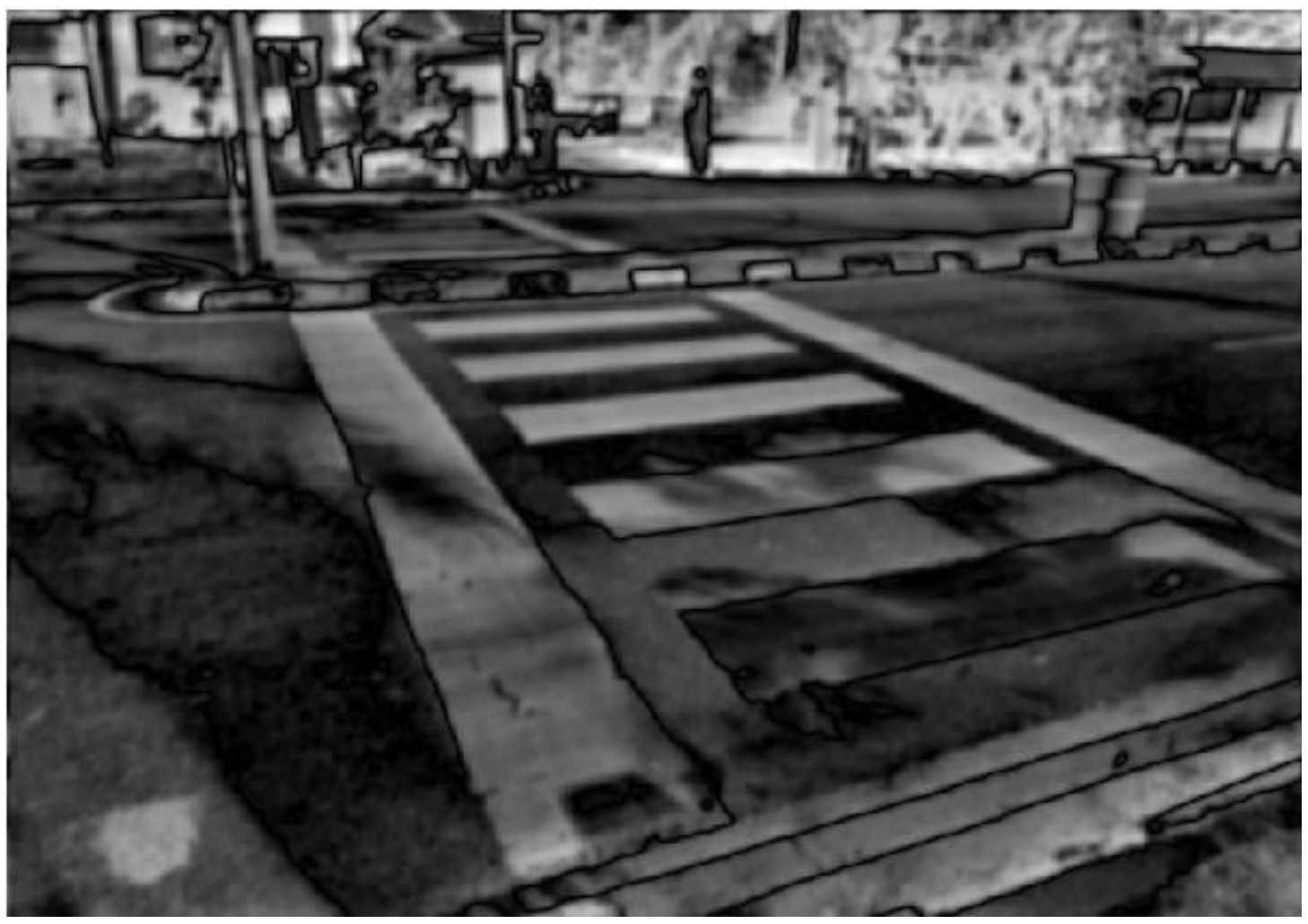

Figure 6. The Wavelet transformation with the "MexicanHat" template is employed to detect the sidewalk.

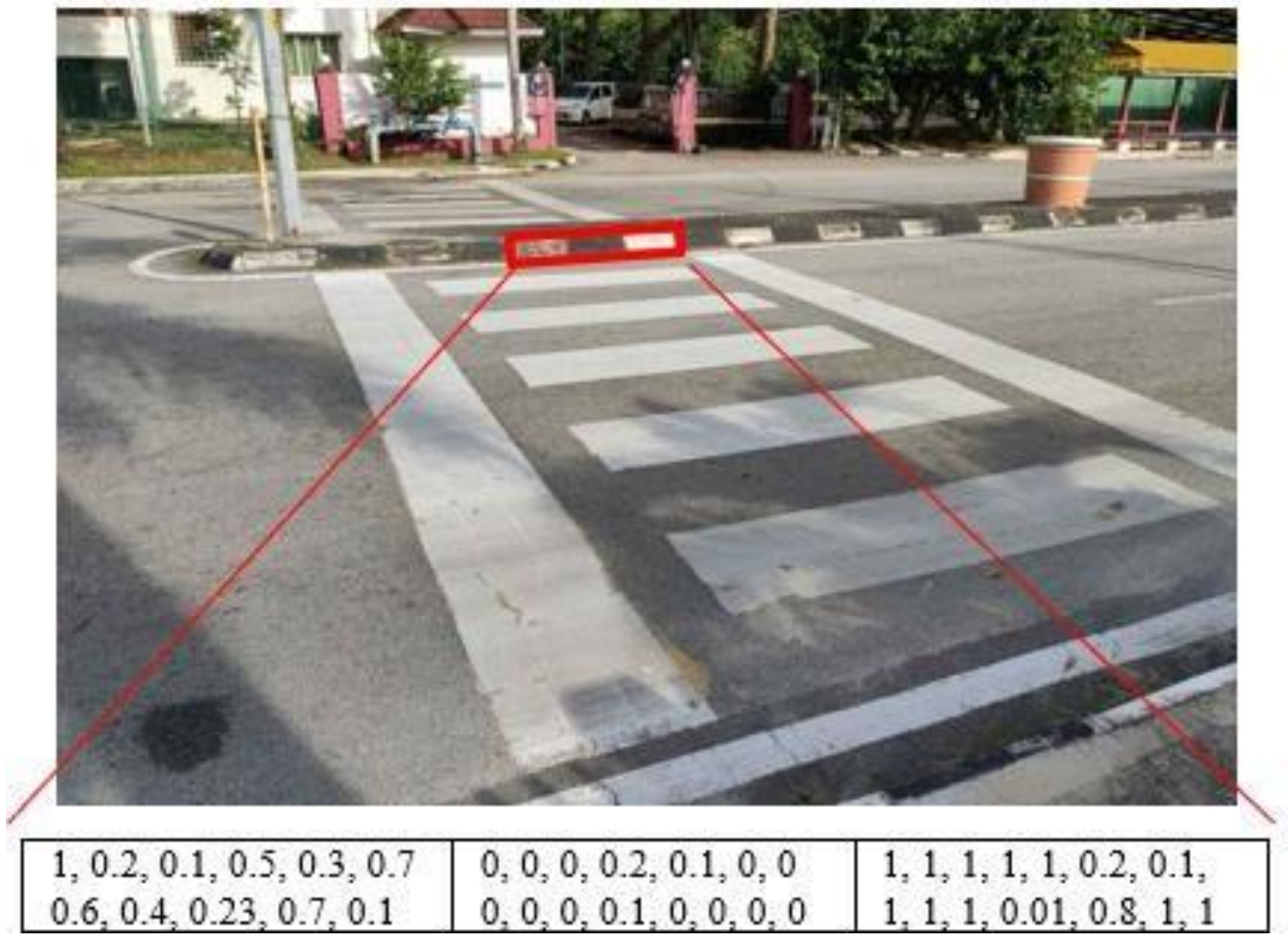

Figure 7. Vague, Dark, and Bright blocks are shown in the illustrated segment. The three blocks will be converted into the vector [0, -1, 1].

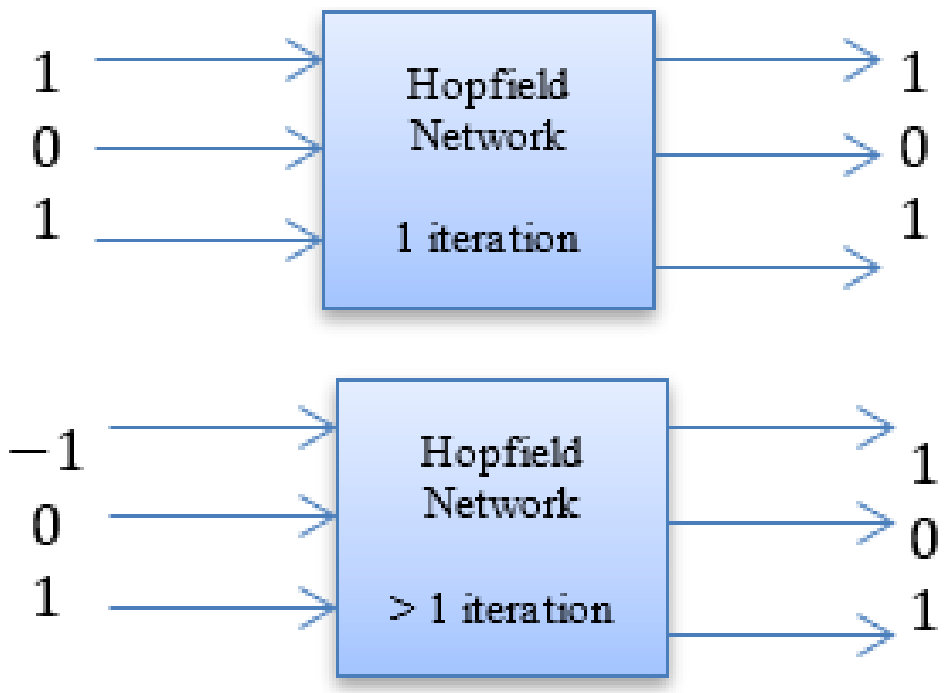

Figure 8. Attracting the nearest vertex requires more than one iteration if the input vector is partially incorrect. In such a case, it will be inferred that a block of this segment should be colored.

## 5 Conclusion:

The importance of unmanned aerial vehicles working as autonomous systems is notably increasing in various civil and military fields. This paper presents a group of recently designed flying robots able to operate in different applications that are aligned with the future of smart cities. Then, we clarify a brief of the technical specifications of such designs, considering their most important elements like sensors and actuators. The inference of smart decisions of intelligent systems relies on a variety of classical and ANN-based computer vision techniques to perform several applications like evaluating the quantity of pesticide that should be applied to infected crops depending on the density of green areas appearing in the given scene, detecting the partially erased blocks of sidewalks that should be colored, or identifying failures in an electrical transformation system. We also spotlight some important details related to the selection of neurons' activation functions during the ANN training process.

## References

[1] Feroz, S., & Dabous, S.A. (2021). UAV-Based Remote Sensing Applications for Bridge Condition Assessment. Remote. Sens., 13, 1809.

[2] Martinez-Alpiste, I., Golcarenarenji, G., Wang, Q., & Calero, J.M. (2021). Search and rescue operation using UAVs: A case study. Expert Syst. Appl., 178, 114937.

[3] Yuan, C., Liu, Z., & Zhang, Y. (2016). Vision-based forest fire detection in aerial images for firefighting using UAVs. 2016 International Conference on Unmanned Aircraft Systems (ICUAS), 1200-1205.

[4] Peng, Z., Zhu, G., Li, M., Zeng, R., Cheng, S., & Wang, K. (2022). A mathematical model for balancing safety and economy of UAVs in forest firefighting. Other Conferences.

[5] Khoufi, I., Laouiti, A., Adjih, C., & Hadded, M. (2021). UAVs Trajectory Optimization for Data Pick Up and Delivery with Time Window. Drones, 5, 27.

[6] Yoon, H., Widdowson, C., Marinho, T., Wang, R.F., & Hovakimyan, N. (2019). Socially Aware Path Planning for a Flying Robot in Close Proximity of Humans. ACM Transactions on Cyber-Physical Systems, 3, 1 - 24.

[7] Suab, S.A., & Avtar, R. (2019). Unmanned Aerial Vehicle System (UAVS) Applications in Forestry and Plantation Operations: Experiences in Sabah and Sarawak, Malaysian Borneo.

[8] Mozaffari, M., Saad, W., Bennis, M., Nam, Y., & Debbah, M. (2018). A Tutorial on UAVs for Wireless Networks: Applications, Challenges, and Open Problems. IEEE Communications Surveys & Tutorials, 21, 2334-2360.

[9] Ghazal, T.M. (2021). RETRACTED ARTICLE: Positioning of UAV Base Stations Using 5G and Beyond Networks for IoMT Applications. Arabian Journal for Science and Engineering, 1 - 1.

[10] Gesbert, P.D. (2020). AUTONOMOUS FLYING ROBOTS for INTELLIGENT IOT DATA HARVESTING.

[11] Giordan, D., Adams, M.S., Aicardi, I., Alicandro, M., Allasia, P., Baldo, M., De Berardinis, P., Dominici, D., Godone, D., Hobbs, P., Lechner, V., Niedzielski, T., Piras, M., Rotilio, M., Salvini, R., Segor, V., Sotier, B., & Troilo, F. (2020). The use of unmanned aerial vehicles (UAVs) for engineering geology applications. Bulletin of Engineering Geology and the Environment, 79, 3437 - 3481.

[12] Radiansyah, S., Kusrini, M.D., & Prasetyo, L.B. (2017). Quadcopter applications for wildlife monitoring. IOP Conference Series: Earth and Environmental Science, 54.

[13] Baba, D. (2020). A new design of a flying robot, with advanced computer vision techniques to perform self-maintenance of smart grids. J. King Saud Univ. Comput. Inf. Sci., 34, 2252-2261.

[14] Nyo, M.T., Mebarek-oudina, F., Hlaing, S.S., & Khan, N.A. (2022). Otsu's thresholding technique for MRI image brain tumor segmentation. Multimedia Tools and Applications, 81, 43837 - 43849.

[15] Baba, D. (2015). Fuzzy Logic-Based Pesticide Sprayer for Smart Agricultural Drone. Journal of Applied Sciences 7(11) (2015).

[16] Baba, D. (2021). Advanced AI-based techniques to predict daily energy consumption: A case study. Expert Syst. Appl., 184, 115508.

[17] Baba, D., Alothman, B. (2023). A fuzzy logic-based stabilization system for a flying robot, with an embedded energy harvester and a visual decision-making system.

[18] Ghiet, A.M.A., Baba, A. (2017). Robot arm control with Arduino. ResearchGate URL: https://doi.org/10.13140/RG.2.2.10227. 53286, doi:10.13140/RG.2.2.10227.53286.